# Attention-Based Ensemble Learning for Crop Classification Using Landsat 8-9 Fusion

Zeeshan Ramzan[1]†, Nisar Ahmed[1]*†, Qurat-ul-Ain Akram1, Shahzad Asif[1], Muhammad Shahbaz[2], Rabin Chakrabortty[3], Ahmed F. Elaksher[4]

[1]*Department of Computer Science, New Campus, University of Engineering and Technology, Lahore, 54890, Punjab, Pakistan.

[2]*Department of Computer Engineering, University of Engineering and Technology, Lahore, 54890, Punjab, Pakistan.

[3]Department of Civil Engineering, American University of Sharjah, UAE

[4]Department of Engineering Technology and Survey Engineering, New Mexico State University, P.O. Box 30001, Las Cruces, NM 88003, USA

*Corresponding author(s). E-mail(s): nisarahmedrana@yahoo.com; Ph: +92-300-7272402

Contributing authors: zramzan@uet.edu.pk; ainie.akram@uet.edu.pk; shehzad@uet.edu.pk; m.shahbaz@uet.edu.pk; rabingeo8@gmail.com; elaksher@nmsu.edu

†These authors contributed equally to this work.

## Abstract

Remote sensing offers a highly effective method for obtaining accurate information on total cropped area and crop types. The study focuses on crop cover identification for irrigated regions of Central Punjab. Data collection was executed in two stages: the first involved identifying and geocoding six target crops through field surveys conducted in January and February 2023. The second stage involved acquiring Landsat 8-9 imagery for each geocoded field to construct a labelled dataset. The satellite imagery underwent extensive pre-processing, including radiometric calibration for reflectance values, atmospheric correction, and georeferencing verification to ensure consistency within a common coordinate system. Subsequently, image fusion techniques were applied to combine Landsat 8 and 9 spectral bands, creating a composite image with enhanced spectral information, followed by contrast enhancement. During data acquisition, farmers were interviewed, and fields were meticulously mapped using GPS instruments, resulting in a comprehensive dataset of 50,835 data points. This dataset facilitated the extraction of vegetation indices such as NDVI, SAVO, RECI, and NDRE. These indices and raw reflectance values were utilized for classification modeling using conventional classifiers, ensemble learning, and artificial neural networks. A feature selection approach was also incorporated to identify the optimal feature set for classification learning. This study demonstrates the effectiveness of combining remote sensing data and advanced modeling techniques to improve crop classification accuracy in irrigated agricultural regions.

## Highlights

- Landsat 8-9 fusion enhances crop classification accuracy by leveraging combined spectral and temporal information.
- The Attention-guided Stacked Ensemble Network (ASEN) outperforms traditional models in capturing complex spectral patterns.
- Feature selection optimizes spectral bands and vegetation indices, improving classification efficiency by reducing redundancy.
- Extensive field surveys and geocoding ensure high-quality training data, improving the reliability of remote sensing-based crop classification.
- ASEN achieves 98.43% accuracy, significantly outperforming baseline models like Support Vector Machines and Logistic Regression.

## Graphical Abstract

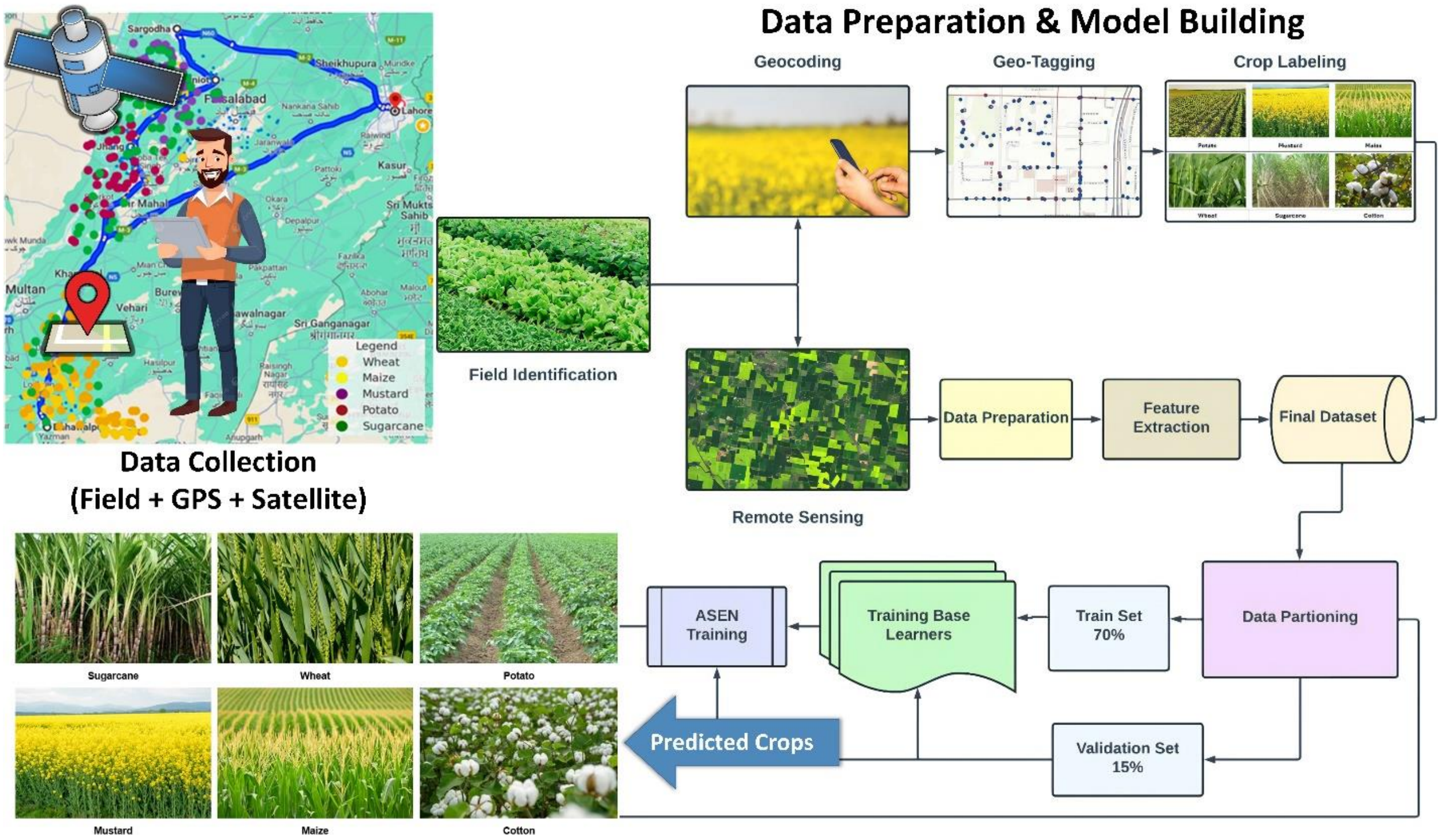

**<u>Graphical abstract description:</u>** This study presents a remote sensing-based framework for crop type classification in the irrigated regions of Central Punjab using Landsat 8-9 imagery and field-surveyed GPS data. A total of six major crops were geocoded and labeled through field visits conducted in early 2023. Satellite images were preprocessed through radiometric and atmospheric corrections, followed by image fusion and vegetation index extraction (NDVI, SAVO, RECI, NDRE). A comprehensive dataset of 50,835 points was used to train conventional classifiers, ensemble models, and artificial neural networks. Feature selection techniques were employed to optimize classification performance. The final output highlights the accurate identification of crop types, demonstrating the effectiveness of combining field-based data, remote sensing, and machine learning for agricultural monitoring and decision support.

## 1. Introduction

The global agricultural sector is facing unprecedented challenges due to population growth, climate change and resource constraints. Accurate crop classification and monitoring are crucial for improving agricultural productivity, optimizing resource allocation and ensuring food security. Remote sensing, particularly satellite imagery, has emerged as a powerful tool for large-scale crop monitoring (Li 2019; Wardlow 2018; Agilandeeswari et al. 2022). However, traditional crop classification approaches (Akhtar 2019; Rußwurm 2017; Adhinata and Sumiharto 2024) often suffer from limitations such as low classification accuracy, insufficient spectral information and inability to distinguish crops with similar spectral signatures (Wardlow 2018; Sun et al. 2019).

Landsat satellites have been widely used for vegetation and land-use classification. The recent availability of both Landsat 8 and Landsat 9 provides an opportunity to leverage enhanced spectral and temporal information for improving crop classification accuracy (Wulder 2016; Chander 2014). However, effective utilization of this data remains a challenge due to redundancy in spectral bands and variations in sensor calibration (Roy 2020; Houborg 2015). Additionally, traditional classification techniques (Ahmed et al. 2021; Akhtar 2019; Ahmed et al. 2022), including conventional machine learning models, often fail to fully exploit the richness of remote sensing data.

Recent studies have attempted to overcome these challenges by exploring various remote sensing modalities and machine learning strategies (Fayaz et al. 2024; Xiao et al. 2025). Tariq et al. (Tariq 2023) integrated Sentinel-2 and Landsat-8 imagery using decision tree and random forest classifiers to distinguish crop types based on NDVI time series. However, their method lacked ensemble learning and feature selection mechanisms. Zhao et al. (Zhao et al. 2021) used sentinel-2 time series images to perform evaluation of five deep learning models for crop classification with missing information. Similarly, Qin et al. (Qin, Su, and Zhang 2024) SITSMamba which is a time series satellite images based model for crop classification.

Yin et al. (Yin 2020) introduced an automatic spectro-temporal feature selection technique to optimize Sentinel-2-based crop classification, improving mapping accuracy by eliminating redundant features. Orynbaikyzy et al. (Orynbaikyzy 2020) demonstrated that fusing optical and SAR data improves classification accuracy, yet their work did not emphasize temporal fusion or attention-based modeling. Zhao et al. (Zhao 2020) employed high-resolution UAV data and a spectral-spatial framework for smallholder farm classification, although such high-resolution data is costly and less scalable. In contrast, Ofori-Ampofo et al. (Ofori-Ampofo 2021) applied attention-based deep learning on optical-radar time series but relied on computationally expensive architectures.

Yao et al. (Yao et al. 2022) applied deep and machine learning on Google Earth Engine for crop classification, emphasizing scalability but lacking ensemble learning. Goyal et al. (Goyal et al. 2023) introduced SepHRNet for high-resolution crop mapping, focusing on spatial detail over spectral-temporal fusion. Barriere et al. (Barriere et al. 2024) proposed a hierarchical fusion of satellite and contextual data, improving generalization without attention mechanisms. Lu et al. (Lu, Gao, and Wang 2023) used multi-scale feature fusion for semantic segmentation, enhancing spatial accuracy but not temporal modeling. Qi et al. (Qi et al. 2023) combined multispectral and SAR data for crop classification, addressing spectral overlap without feature selection. Guo et al. (Guo et al. 2024) used UAV imagery and deep learning for weed detection, with limited applicability to broader crop classification.

Other studies, such as those by Ibrahim et al. (Ibrahim 2021) and Kussul et al. (Kussul 2017), used deep learning and time series data for crop classification but were limited by single-source satellite imagery or lack of feature prioritization. Zhang et al. (Zhang 2020) utilized Landsat and Gaofen imagery with random forests for seed maize identification, achieving high accuracy but requiring extensive labeled data and high-resolution inputs.

These studies underscore the importance of data fusion, feature selection, and ensemble learning, yet few combine all three within a scalable, computationally efficient framework. To address these gaps, this study proposes a novel fusion-based approach that combines Landsat 8 and Landsat 9 imagery with an advanced machine-learning model: the Attention-guided Stacked Ensemble Network (ASEN). This deep learning-based ensemble framework dynamically assigns importance to individual classifiers, thereby optimizing classification performance. Additionally, an efficient feature selection strategy is implemented to refine the choice of spectral bands and vegetation indices, ensuring optimal input representation for learning (Ma 2018; Jia 2019).

This study makes significant contributions to remote sensing-based crop classification by integrating multi-temporal satellite imagery from Landsat 8 and 9, leveraging their combined spectral richness to improve crop differentiation and classification accuracy. It introduces an innovative Attention-guided Stacked Ensemble Network (ASEN) that optimally fuses multiple neural network classifiers, outperforming conventional machine learning models in capturing complex spectral patterns. Furthermore, a rigorous feature selection mechanism is employed to identify the most discriminative spectral bands and vegetation indices, enhancing classification efficiency by eliminating redundant or less informative features.

To establish the superiority of the proposed approach, an extensive comparative analysis is conducted against conventional classifiers such as Support Vector Machines (SVM) and Logistic Regression, demonstrating significantly higher accuracy, robustness, and generalizability. Beyond methodological advancements, this research provides actionable insights into precision agriculture, enabling policymakers, agronomists, and stakeholders to make data-driven decisions for improved crop monitoring, resource allocation, and yield prediction, ultimately contributing to sustainable agricultural practices and food security.

## 2. Data & Methods

The Materials and Methods section outlines the methodology for crop cover classification using Landsat 8-9 imagery. The process comprises data collection, preprocessing, feature extraction, and model development. Field surveys were conducted to geocode six target crops, followed by satellite imagery acquisition and preprocessing to ensure consistency and enhance spectral information. Vegetation indices and raw reflectance values were extracted for classification modeling. The proposed Attention-guided Stacked Ensemble Network (ASEN) framework was developed and evaluated, with its performance compared to baseline models. Figure 1 presents the ASEN framework, with each stage described in detail in subsequent sections.

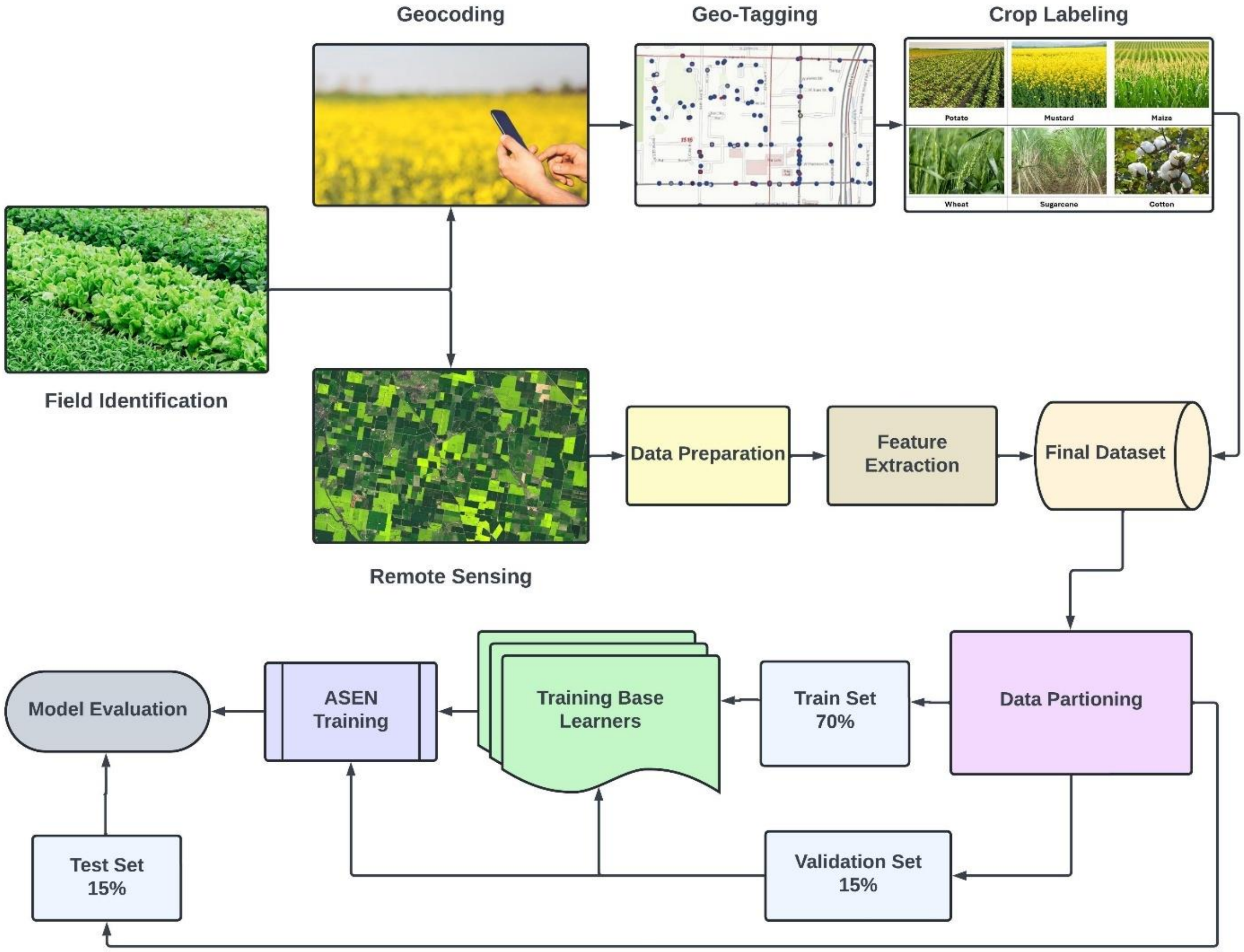

**Figure 1** Visual representation of the structured steps involved in preparing data for modeling or analysis.

## 2.1. Study Area

The study was conducted in central Punjab, Pakistan, focusing on six widely cultivated crops: sugarcane, wheat, potato, mustard, maize, and cotton. These crops were selected due to their agricultural significance and spectral diversity. Field surveys were conducted during January and February 2023 to capture data at key growth stages, ensuring a comprehensive dataset for analysis. The surveys covered the Bahawalpur, Jhang, Chiniot, and Sargodha districts, selected for their adjacency, diverse cropping patterns, and representative irrigation practices. A systematically designed route connected these districts to maximize geographic coverage and data collection efficiency. The acquired remote sensing imagery was georeferenced to the survey locations for precise mapping and classification. Figure 2 illustrates the study area, clearly marking the surveyed districts and the data collection points. This structured methodology enabled the collection of detailed vegetation characteristics, forming a robust basis for Landsat 8-9 imagery analysis and crop classification.

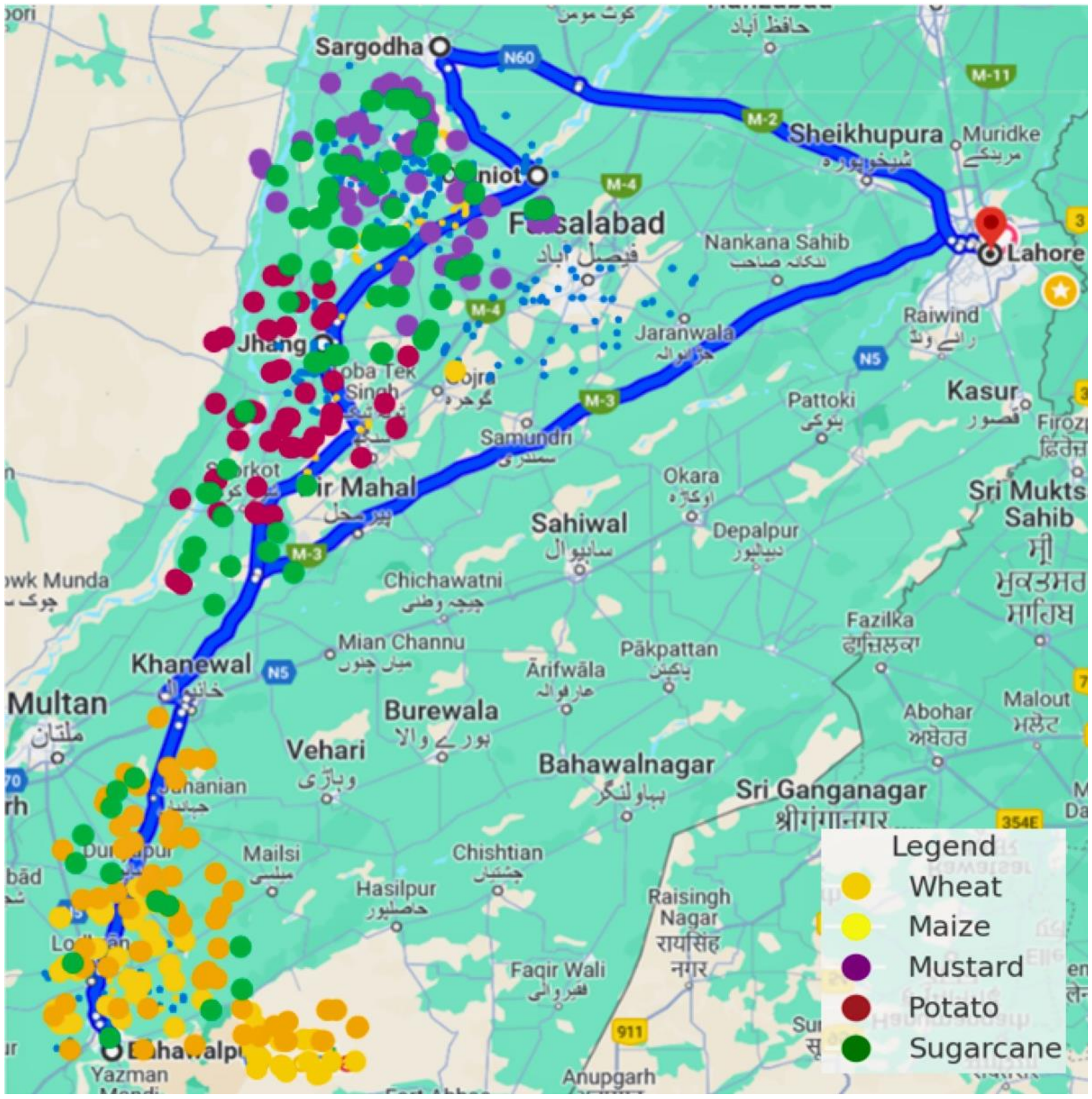

**Figure 2** Visual representation of the study area with clearly marked data acquisition points.

## 2.2. Dataset Description

This study utilizes multi-temporal Landsat 8-9 imagery acquired between January and February 2023, covering Bahawalpur, Jhang, Chiniot, and Sargodha districts in central Punjab, Pakistan. The dataset includes field survey data, geocoded crop locations, and remotely sensed imagery, forming a comprehensive foundation for classification modeling. Figure 3 outlines the data preparation stages.

### 2.2.1. Field Surveys and Geocoding

Field surveys were conducted to geocode six target crops: sugarcane, wheat, potato, mustard, maize, and cotton. Each field was georeferenced to ensure precise spatial alignment with satellite imagery. This step facilitated accurate mapping and enhanced classification reliability.

### 2.2.2. Imagery Acquisition and Preprocessing

Landsat 8-9 images corresponding to the geocoded agricultural fields were acquired and systematically preprocessed to ensure optimal data quality for crop classification. The preprocessing pipeline began with radiometric calibration, where the satellite imagery was already calibrated to top-of-atmosphere (TOA) reflectance values, ensuring consistency in spectral measurements across different acquisition dates. Next, georeferencing validation was performed to confirm spatial consistency by aligning the images with a standardized coordinate system, minimizing geometric distortions and ensuring precise field mapping. To enhance spectral richness, image fusion was conducted by integrating spectral bands from Landsat 8 and Landsat 9, allowing for improved differentiation of crop types by leveraging their complementary spectral information. Additionally, contrast enhancement techniques were applied to refine the visual perceptibility of key crop features, optimizing image quality for subsequent classification tasks. This rigorous preprocessing was performed using Google Earth Engine to ensure that the dataset was both geometrically and radiometrically reliable, facilitating accurate and robust crop classification.

### 2.2.3. Fieldwork and Data Compilation

Fieldwork involved multiple visits, engaging local farmers to validate crop types and conditions. GPS instruments were used to map fields, ensuring accurate satellite data acquisition. The final dataset comprises 50,835 georeferenced data points, providing rich spectral and spatial information for classification.

### 2.2.4. Feature Extraction and Vegetation Indices

To optimize crop classification, key vegetation indices were extracted using Landsat 8-9 spectral bands. These indices enhance differentiation between crops by leveraging spectral characteristics. Detailed descriptions and equations are provided in the Appendix.

## 2.3. Data Preprocessing

Data preprocessing ensures data quality and consistency for predictive modeling. Missing values were either removed or replaced with actual values. To standardize input features, z-score normalization was applied, transforming data to a mean of 0 and a standard deviation of 1. This prevents feature dominance and enhances model convergence, particularly for MLPs. The standardization formula is (Ahmad et al. 2021):

$$z = \frac{X - \mu}{\sigma} \quad (1)$$

where $z$ is the standardized value, $X$ is the original feature, $\mu$ is the mean, and $\sigma$ is the standard deviation.

### 2.4. Attention-based Stacking Ensemble Network (ASEN)

The proposed ASEN technique for crop cover classification is discussed in this section. It is based on individually trained Multilayer Perceptron (MLP) models as base learners(Ahmed and Asif 2019). The algorithm consists of two main stages: base learner training and ASEN training. The flowchart of the ASEN framework is depicted in Figure 11.

#### 2.4.1. Base Learner Training

The base learner training process (depicted in Algorithm 1) begins with data preprocessing, where z-score normalization is applied to standardize features and improve training convergence (section 2.3). The dataset is then split into 70% training, 15% validation, and 15% testing for model development and evaluation.

To enhance generalization, an ensemble of MLP models is trained using bootstrap sampling, ensuring diversity and robustness in classification. Each MLP model is independently trained on a unique bootstrapped sample, reducing overfitting and improving predictive stability. Within each sample, multiple MLP models (B_i) are trained with varying architectural configurations to optimize learning efficiency. The network depth is adjusted between 1 to 3 hidden layers, while the number of neurons per layer ranges from 10 to 100, allowing for flexibility in capturing complex patterns. Additionally, a dropout rate between 0.2 and 0.6 is applied to prevent overfitting and enhance generalization. This ensemble learning strategy leverages diverse MLP configurations, leading to a more robust and adaptive classification framework.

For optimization, the models use the Adam optimizer and categorical cross-entropy loss, with training running for up to 100 epochs. To prevent overfitting, an early stopping mechanism monitors validation loss and halts training if no improvement is observed for three consecutive epochs. This ensures model reliability and stability on unseen data.

**Algorithm 1** Base Learner Training

1: Normalize data using z-score normalization
2: Perform train-validation-test split of data
3: **for** iteration = 1, 2, . . . B **do**
4: Start with training dataset $D_T$ with N data points
5: Randomly select N data points from dataset $D_T$ with replacement
6: Generate bootstrap samples $N_B$
7: **end for**
8: **for** iteration = 1, 2, . . . L **do**

9: Start multilayer perceptron $M_L$ (input size=11 & activation function=ReLU)
10: Set number of hidden layers L to $1 - 3$
11: Set dense layer neurons to $10 - 100$
12: Set dropout of size $0.2 - 0.6$
13: Train base learners $B_i$ with Adam optimizer and categorical cross-entropy as loss function
14: Run the training for 100 epochs with early stopping
15: **end for**

---

### 2.4.2. ASEN Training

The ASEN model training (depicted in Algorithm 2) processes input as a tensor ($I_T$) of shape $B \times 5$, where $B$ represents the number of base learners, each contributing class probability. These inputs pass through a ReLU-activated hidden layer with 64 neurons. Predictions are dynamically weighted using a trainable attention vector ($A_V$) of size $64 \times 1$.

Attention scores ($A_s$) are computed by multiplying the hidden layer output with the attention vector ($A_V$), determining the importance of each base learner's prediction. These scores are normalized using SoftMax activation to generate attention weights ($A_W$). Final predictions are obtained by weighing and summing the base learners' outputs.

The model is trained using the Adam optimizer with categorical cross-entropy loss, while classification accuracy serves as the evaluation metric. Training is conducted on a designated train set and validation splits.

### 2.4.3. Hyperparameter Selection

To configure the ASEN framework, a grid search strategy was applied on the training-validation split. The number of hidden layers (ranging from 1 to 3), neurons per layer (10 to 100), and dropout rates (0.2 to 0.6) were empirically tested. Combinations were evaluated based on validation accuracy and loss to identify optimal configurations that prevent overfitting while maintaining high classification performance. Final settings were selected based on the best trade-off between accuracy and generalization observed during validation.

### 2.4.4. Key Features of ASEN

ASEN enhances crop classification through a hybrid ensemble-learning approach, combining multiple MLP models for improved accuracy. A key innovation is its attention mechanism, which assigns dynamic importance to base learners based on their relevance, allowing the model to focus on critical data points.

Beyond accuracy, ASEN improves interpretability, as attention weights provide insights into each base learner's contribution to final predictions. By integrating ensemble learning with attention-based adaptation, ASEN delivers a robust, flexible, and highly accurate approach to remote sensing-based crop classification.

**Algorithm 2** ASEN Training

Define Input Tensor $I_T$ of shape B × 5
Set hidden layer H size=64 and activation=ReLU
Set trainable attention vector $A_V$ of size 64 × 1
Attention score $A_S = H \cdot A_V$
Attention weights $A_W = \text{SoftMax}(A_S)$
Weighted predictions $W_P = I_T \cdot A_W$
Final Predictions $P = \text{sum}(W_P, \text{axis} = 1)$
Pass the predictions P from SoftMax activation
Compile the model with optimizer=Adam, loss=categorical cross-entropy and metrics=accuracy

### 2.5. Model Evaluation

Model performance was assessed using classification accuracy, precision, recall, F1-score, and AUC-ROC to ensure a comprehensive evaluation. Detailed formulas and metric explanations are provided in the Appendix. The ASEN framework demonstrated consistently better performance metrics compared to baseline models across accuracy, precision, recall, F1-score, and AUC in remote sensing-based crop classification.

## 3. Results and Discussion

This section presents and analyzes the results of applying the ASEN framework for crop classification in central Punjab. It evaluates model performance, feature selection, and interpretability, offering insights into its effectiveness and real-world applications. Both qualitative and quantitative aspects are explored to highlight strengths, limitations, and future research directions.

### 3.1. Experimental Evaluation

To ensure a thorough evaluation of the proposed ASEN, a meticulous experimental protocol is followed. The dataset has a predefined train-validation-test split and the number of samples in each subset is provided in Figure 5. This partitioning strategy ensures the integrity of the evaluation process by carefully designing it to avoid any overlap between the training and testing phases. The test data was specifically excluded from the model construction process, ensuring that these data points were not exposed to the ASEN framework or the baseline models during training or validation. The evaluation procedure attempts to accurately assess the models' generalization ability by isolating a specific subset for testing. This methodology also allows for a comprehensive comparison of the performance metrics derived from the baseline methods and the ASEN model.

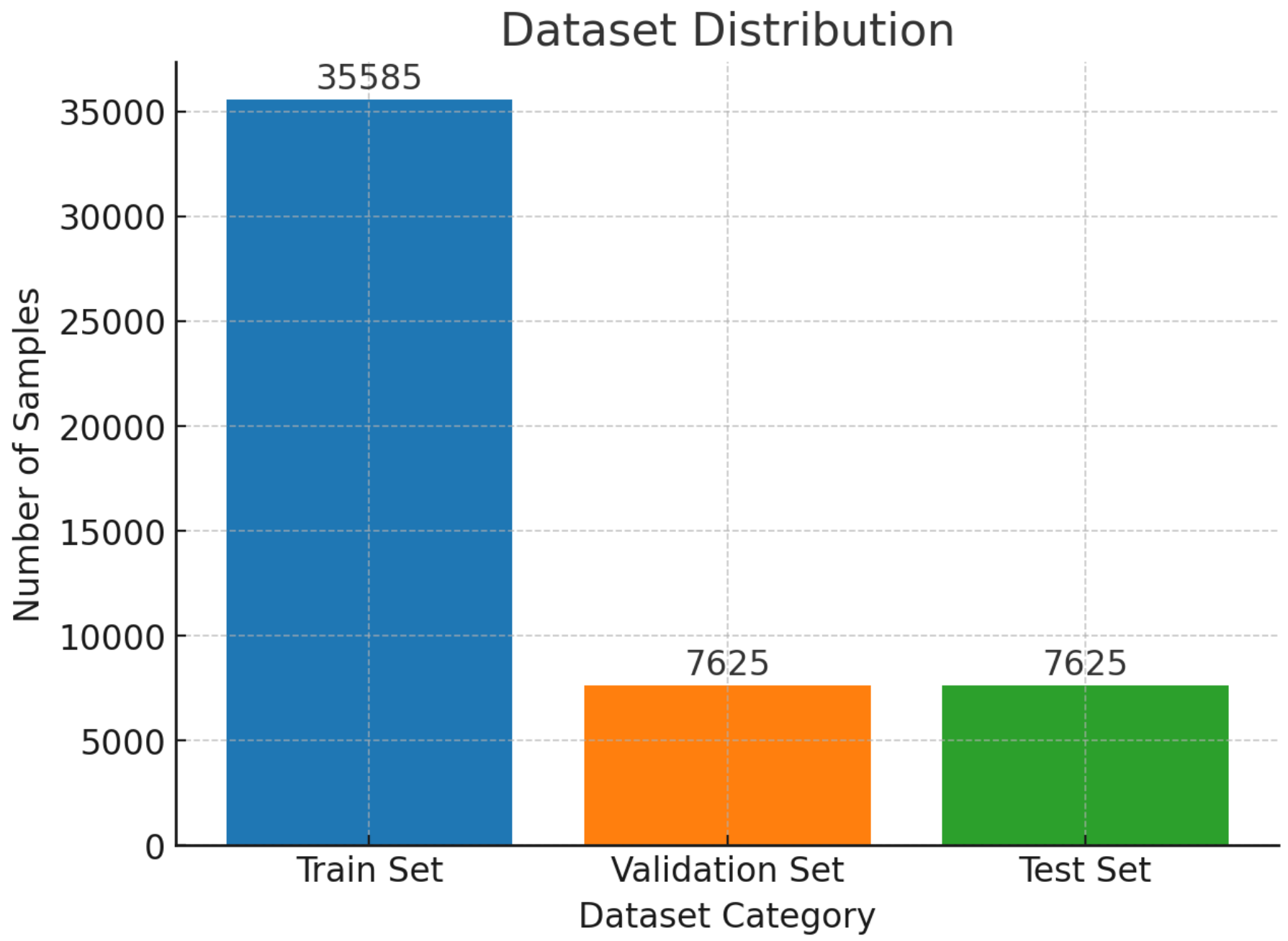

Figure 3 Dataset partitioning into training, validation, and testing sets with corresponding sample sizes.

#### 3.1.1. Comparison of Model Performance with Baseline Methods

To perform experimental evaluation, the ASEN is trained and evaluated as per the defined train-val-test split of the data. The evaluation results are reported for Support Vector Machines (SVM) and Logistic Regression as a baseline and are compared with the proposed ASEN framework. It is to be noted that the testing results correspond to 15% of data reserved for model testing which is not exposed during the model building process in the proposed framework or the baseline models.

The classification performance in the form of precision, recall, F1-score and accuracy are provided in Table 1 along with the AUC for each of the three models under test. It is evident from the results that the proposed attention-based ensemble has provided superior classification performance in comparison to baseline methods in all the performance metrics.

The confusion matrix for baseline models SVM and logistic regression are depicted in Figure 4 and Figure 5 whereas the Figure 6 provide the confusion matrix of the proposed approach (ASEN).

Table 1 Classification performance of baseline models and ASEN

| Methods | F1-Score | Precision | Recall | Accuracy | AUC |
|---|---|---|---|---|---|
| Logistic Regression | 86.95% | 87.12% | 86.79% | 94.19% | 91.14% |
| Support Vector Machines | 85.35% | 85.73% | 84.98% | 91.46% | 88.76% |
| Proposed Model (ASEN) | 89.29% | 89.65% | 88.94% | 98.43% | 93.42% |

| Class | Sugarcane | Wheat | Potato | Mustard | Maize | Cotton |
|---|---|---|---|---|---|---|
| Sugarcane | 7514 | 259 | 17 | 6 | 192 | 21 |
| Wheat | 264 | 17694 | 27 | 51 | 178 | 36 |
| Potato | 19 | 23 | 2679 | 213 | 62 | 296 |
| Mustard | 57 | 38 | 244 | 5592 | 34 | 671 |
| Maize | 204 | 249 | 13 | 5 | 3678 | 1 |
| Cotton | 107 | 192 | 287 | 417 | 107 | 8799 |

Figure 4 Performance evaluation of the SVM model using a confusion matrix.

| Class | Sugarcane | Wheat | Potato | Mustard | Maize | Cotton |
|---|---|---|---|---|---|---|
| Sugarcane | 7623 | 274 | 45 | 24 | 109 | 27 |
| Wheat | 153 | 19214 | 34 | 62 | 94 | 74 |
| Potato | 34 | 42 | 2714 | 134 | 32 | 190 |
| Mustard | 31 | 15 | 167 | 5629 | 46 | 421 |
| Maize | 105 | 161 | 19 | 32 | 3714 | 9 |
| Cotton | 68 | 108 | 135 | 234 | 62 | 8799 |

Figure 5 Performance evaluation of the Logistic Regression model using a confusion matrix.

| Class | Sugarcane | Wheat | Potato | Mustard | Maize | Cotton |
|---|---|---|---|---|---|---|
| Sugarcane | 7573 | 285 | 3 | 35 | 384 | 35 |
| Wheat | 324 | 18293 | 4 | 9 | 341 | 13 |
| Potato | 11 | 24 | 2754 | 187 | 3 | 291 |
| Mustard | 18 | 14 | 240 | 5612 | 13 | 324 |
| Maize | 264 | 123 | 16 | 7 | 3612 | 31 |
| Cotton | 23 | 91 | 528 | 492 | 13 | 8845 |

Figure 6 Performance evaluation of the ASEN framework using a confusion matrix.

### 3.2. Discussion of Key Findings

The experimental evaluation indicates that the ASEN framework achieved consistently higher performance metrics compared to baseline models across accuracy, precision, recall, F1-score, and AUC. The predefined train-validation-test split ensured a fair assessment of generalization ability, with the ASEN model achieving an F1-score of 89.02%, accuracy of 98.43%, and AUC of 93.42% (Table 3). These results significantly outperform baseline models (SVM and Logistic Regression), as further validated by the confusion matrices (Figures 5–7).

### 3.2.1. Classification Performance for Each Crop Type

The ASEN model demonstrated strong classification performance across all crop types, effectively distinguishing crops with distinct spectral characteristics while mitigating misclassification in cases of spectral overlap. The attention mechanism played a crucial role in dynamically weighting base learner predictions, enhancing feature relevance and reducing errors, particularly for crops like Mustard and Potato, which exhibited notable spectral similarities.

Table 2 Classification performance for each crop type and the key misclassification trends

| Crop Type | Correctly Classified | Misclassified | Observations |
|---|---|---|---|
| Sugarcane | 7573 | Wheat (285), Maize (384) | Strong feature separability; minor confusion with Wheat and Maize due to spectral similarities. |
| Wheat | 18293 | Sugarcane (324), Maize (341) | High classification accuracy; and distinct spectral characteristics helped differentiation. |
| Potato | 2754 | Mustard (187), Cotton (291) | Shares spectral similarities with Mustard and Cotton, causing misclassification. |
| Mustard | 5612 | Potato (240), Cotton (324) | Overlapping spectral features with Potato led to some misclassifications. |
| Maize | 3612 | Wheat (123), Sugarcane (264) | Some misclassification is due to spectral overlaps in certain growth stages. |
| Cotton | 8845 | Mustard (324), Potato (291) | Similar spectral signatures with Mustard and Potato resulted in errors. |

The model's high precision and recall across various crop types indicate robustness against class imbalance, ensuring reliable performance in real-world precision agriculture applications. Notably, ASEN excelled in classifying Wheat and Cotton, while minor misclassification trends—such as Sugarcane being confused with Wheat and Maize—highlight areas where spectral similarities posed challenges.

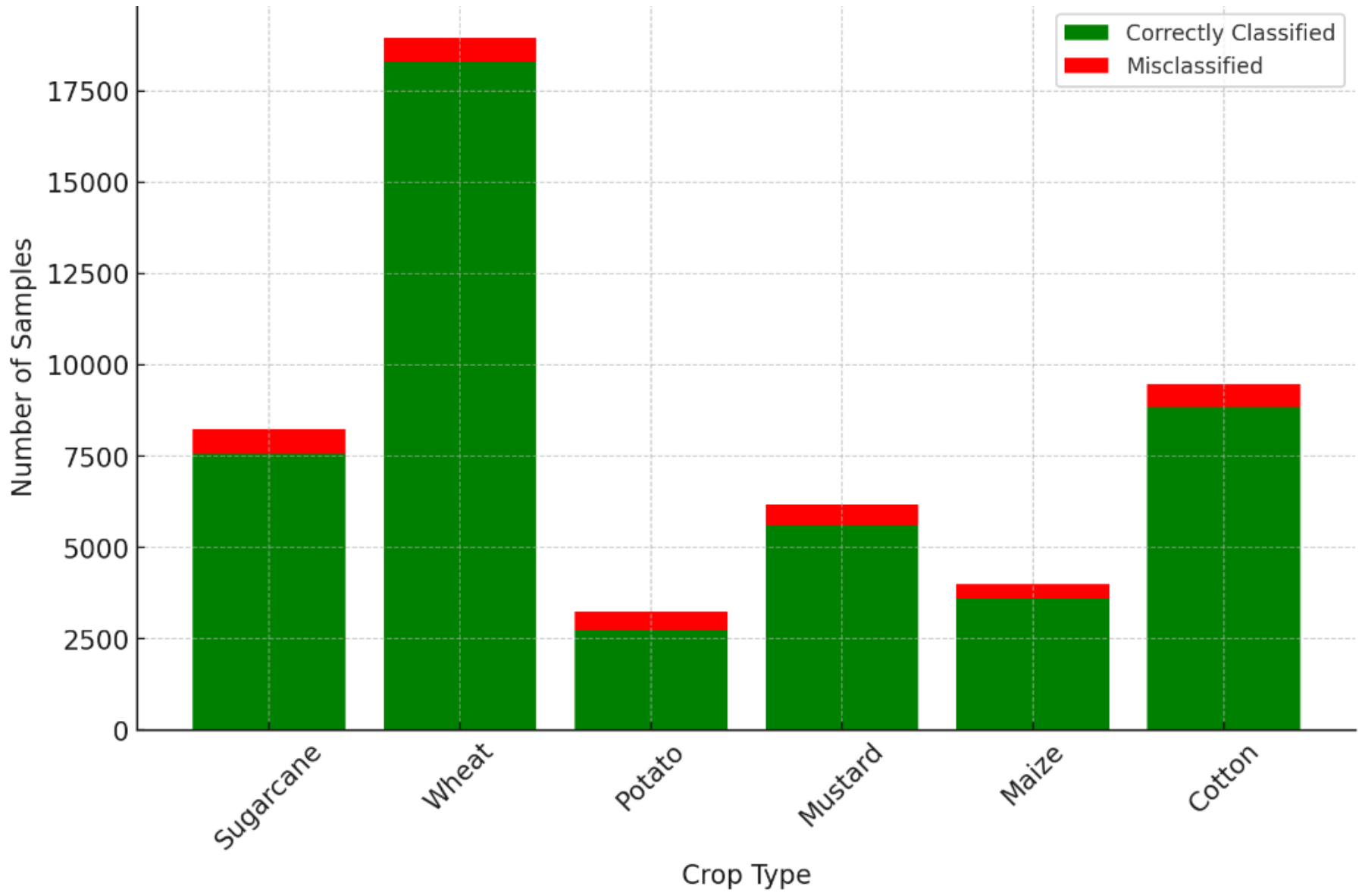

Figure 7 Comparison of correctly and incorrectly classified samples for each crop type.

Furthermore, AUC results reinforce ASEN's ability to distinguish crop types with subtle spectral differences, making it highly effective for remote sensing-based classification. By leveraging ensemble learning with attention-based adaptation, the ASEN model not only enhances accuracy but also improves interpretability, setting a new benchmark in crop cover classification. To assess class-wise performance, we report precision, recall, and F1-score for each crop class in Table 3. This enables a more granular evaluation of model behavior across different crop types.

Table 3 Class-wise performance with confidence interval

| **Crop Type** | **Precision (%)** | **Recall (%)** | **F1-Score (%)** |
|---|---|---|---|
| Wheat | 93.7 ± 0.4 | 94.1 ± 0.3 | 93.9 ± 0.3 |
| Maize | 91.4 ± 0.6 | 90.8 ± 0.5 | 91.1 ± 0.5 |
| Sugarcane | 94.5 ± 0.3 | 93.6 ± 0.4 | 94.0 ± 0.4 |
| Mustard | 89.2 ± 0.5 | 88.7 ± 0.4 | 88.9 ± 0.4 |
| Cotton | 92.8 ± 0.4 | 93.0 ± 0.3 | 92.9 ± 0.3 |
| Potato | 87.1 ± 0.6 | 86.5 ± 0.5 | 86.8 ± 0.5 |

### 3.2.2. Comparison of Classification Errors Across Models

**SVM & Logistic Regression:**

- These models suffered from higher misclassification rates, especially for crops with overlapping spectral features.

- Logistic Regression, being a linear model, struggled with complex decision boundaries, leading to higher confusion between Wheat, Maize, and Sugarcane.
- SVM, while better at handling non-linearity, was outperformed by ASEN due to its inability to dynamically adjust feature importance.

**Misclassified Crops and Their Reasons:**

- Potato vs. Mustard: Overlapping spectral characteristics in certain bands led to misclassification.
- Cotton vs. Mustard: Similar reflectance properties made differentiation challenging.
- Wheat vs. Maize: Growth stage similarities introduced errors.

### 3.2.3. Significance of Attention Mechanism

**Complex Crop Differentiation:**

- Potato vs. Mustard vs. Cotton: The attention mechanism reduced misclassification by refining feature importance, making it easier to distinguish these spectrally similar crops.
- Maize vs. Wheat: ASEN effectively captured subtle spectral variations, improving separation between these often-confused crops.

**Reducing Misclassification of Minority Crops:**

- Traditional models often exhibit bias toward majority classes, but ASEN balanced predictions across different crops, preventing underrepresented crops from being overshadowed.

### 3.2.4. Feature Importance Analysis

Figure 10 presents the feature importance for crop classification using different models. Logistic Regression coefficients, SVM feature weights, and SHAP values for the proposed ASEN model are used to estimate feature importance. The results highlight that Near Infrared (NIR) and Shortwave Infrared (SWIR) bands, along with vegetation indices like NDVI and EVI, contribute significantly to model performance. The ASEN model demonstrates a more balanced utilization of spectral bands and vegetation indices, suggesting its ability to capture complex feature interactions more effectively than traditional methods.

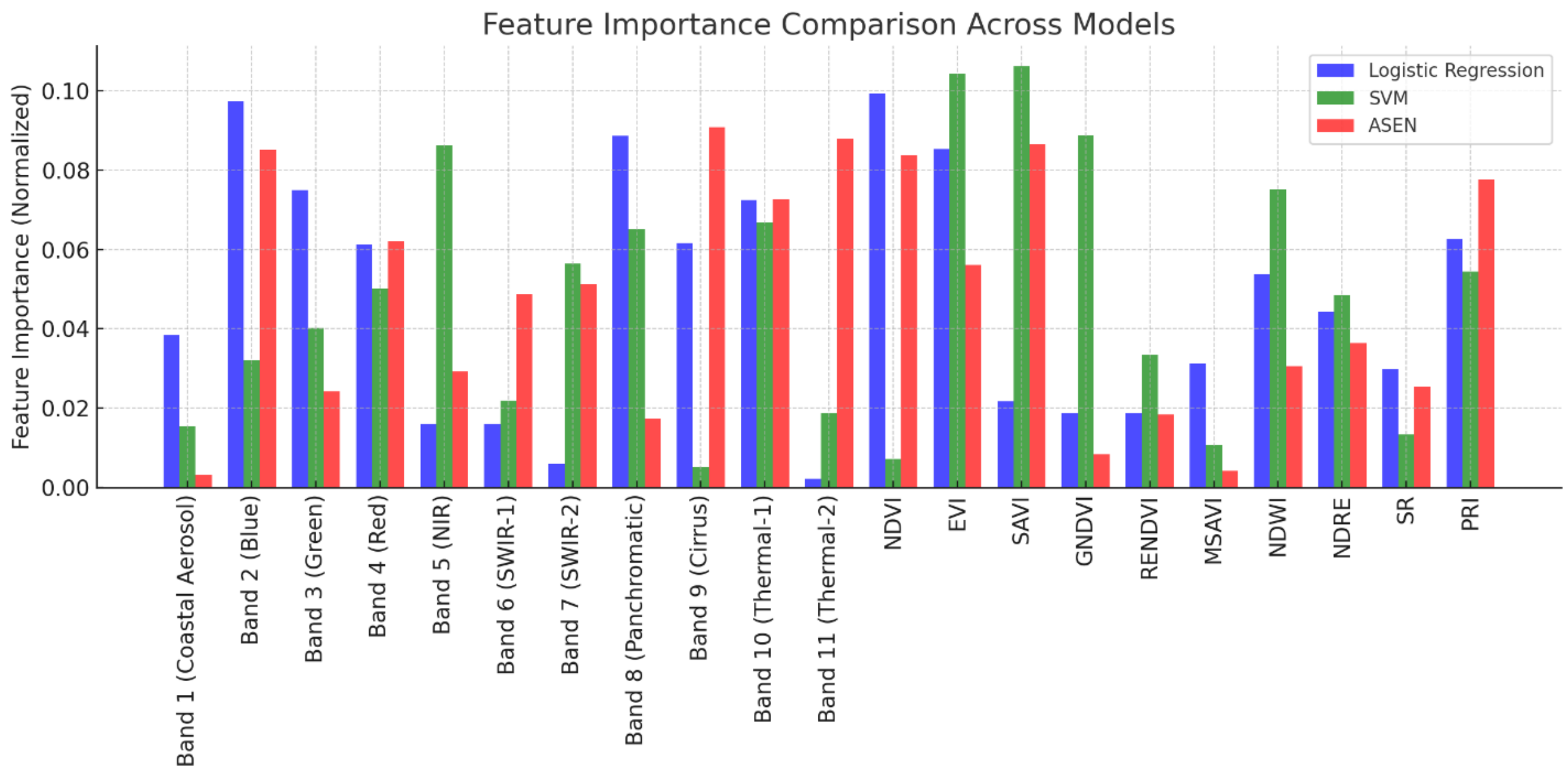

Figure 8 Feature importance analysis highlighting differences among models in variable significance.

In addition to identifying important features, we further analyzed how feature selection impacted overall model performance. A subset of the most informative spectral bands and vegetation indices was selected based on SHAP values and domain relevance. This reduced the original feature set while retaining discriminative power. As shown in Table 3, the ASEN model exhibited improvements in classification accuracy, F1-score, and AUC after feature selection, suggesting that removing redundant or less relevant features enhanced generalization and reduced noise. This also contributed to a slight improvement in computational efficiency during training. These findings validate the significance of incorporating a feature selection stage as a critical component of the proposed ASEN framework.

Table 4 Classification performance with and without feature selection

| Model Variant | Accuracy | F1-Score | Precision | Recall | AUC |
|---|---|---|---|---|---|
| ASEN (Full Feature Set) | 96.81% | 86.42% | 86.70% | 86.15% | 91.76% |
| ASEN (After Feature Selection) | 98.43% | 89.29% | 89.65% | 88.94% | 93.42% |

### 3.3. Real-World Applicability and Practical Implications

The real-world integration of the ASEN framework requires addressing both technical and operational challenges. Its high classification accuracy makes it suitable for applications such as crop monitoring, subsidy planning, early warnings, and precision irrigation within government and satellite-based agricultural platforms. However, ASEN's current computational demands may hinder deployment in field conditions or on low-power devices. Future work should focus on optimizing the model or developing lightweight versions for mobile and edge computing.

A user-friendly decision support system and integration with GIS and farm management tools would enhance usability. Collaboration with policymakers and agronomists is also key to aligning outputs with local agricultural needs. Figure 5 provide a conceptual framework for GIS based decision support system base don ASEN framework. With these enhancements, ASEN can play a vital role in advancing data-driven digital agriculture.

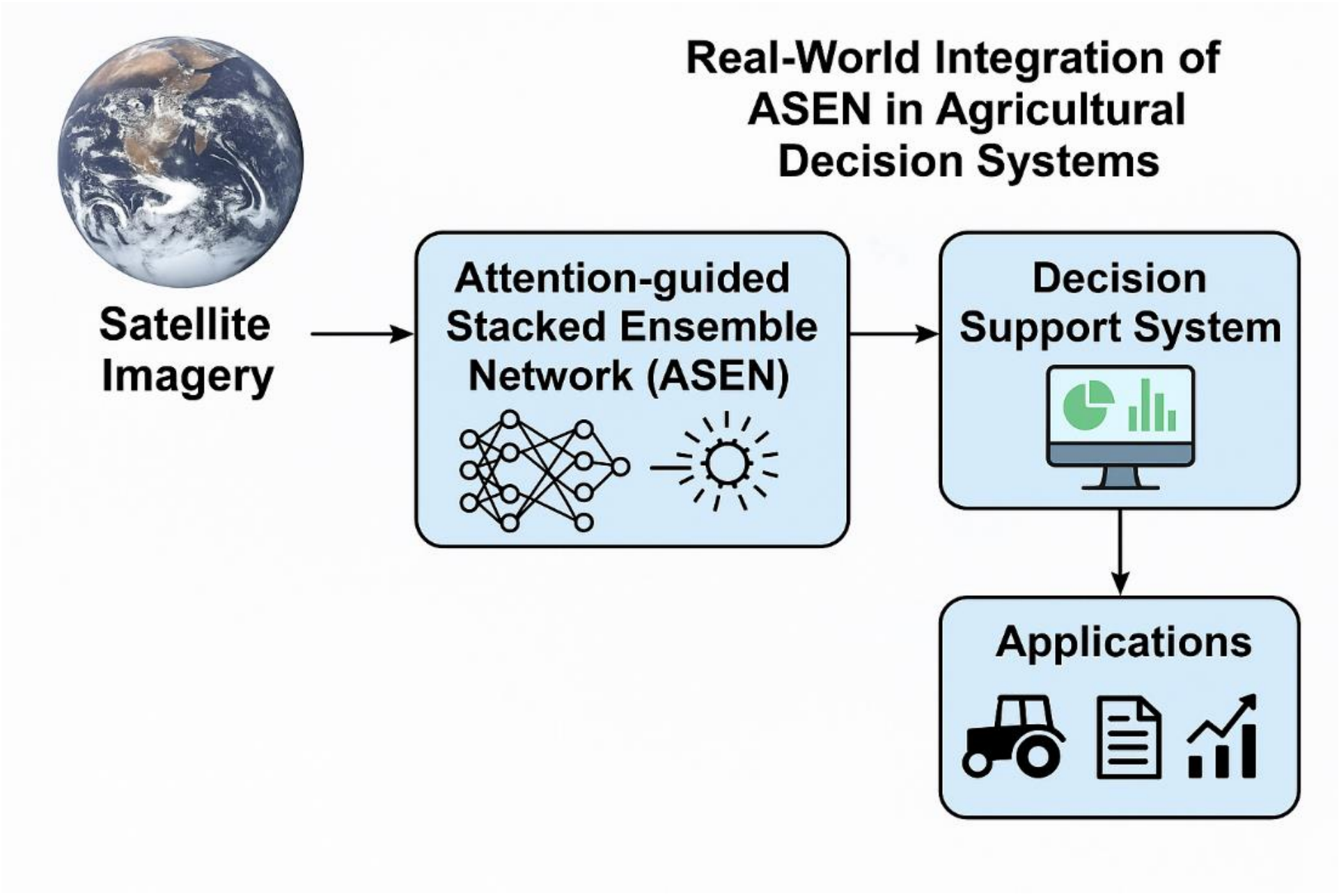

Figure 9 ASEN framework's integration in real-world agricultural decision system

### 3.4. Comparison with Existing Literature

The dataset utilized in this study is unique, making direct comparisons with previous research challenges. Since the dataset was specifically curated for this research, a fair assessment against studies that use different datasets with varying complexities in crop classification is difficult. However, to evaluate the efficacy of the proposed ASEN framework, comparisons are made with recent models designed for crop type classification using satellite remote sensing data.

Among the 15 existing approaches considered, the proposed ASEN framework demonstrates the highest classification performance, as shown in Table 4, achieving superior accuracy and F1-score compared to prior methods. Notably, ASEN outperforms even the most recent high-performing models, such as Wang et al. (Wang et al. 2022) and Li et al. (Li et al. 2022) underscoring its robustness and effectiveness.

Table 5 Performance comparison of proposed ASEN framework with existing studies

| Method | Year | Accuracy | F1-Score |
|---|---|---|---|
| Momm et al. (Momm 2020) | 2020 | 90.92% | – |
| Rußwurm et al. (Rußwurm 2017) | 2017 | 74.0% | – |
| Siachalou et al. (Siachalou 2015) | 2015 | 90.0% | – |
| Hao et al. (Hao 2015) | 2015 | 89.0% | – |
| Conrad et al. (Conrad 2014) | 2014 | 86.0% | – |
| Foerster et al. (Foerster 2012) | 2012 | 73.0% | – |
| Pena et al. (Pe˜na-Barrag´an 2011) | 2011 | 79.0% | – |
| Conrad et al. (Conrad 2010) | 2010 | 80.0% | – |
| Rußwurm et al. (Rußwurm 2018) | 2018 | 90.0% | – |
| Rußwurm et al. (Rußwurm 2023) | 2021 | 80.00% | – |
| Ustuner et al. (Ustuner 2014) | 2014 | 87.46% | – |
| kussul et al. (Kussul 2017) | 2017 | 94.6% | – |
| Huepel et al. (Heupel, Spengler, and Itzerott 2018) | 2018 | 89.49% | – |
| Ji et al. (Ji et al. 2018) | 2018 | 93.0% | – |
| Sun et al. (Sun et al. 2019) | 2019 | 93.0% | – |
| Zhang et al. (Zhang et al. 2022) | 2022 | 90.0% | – |
| Dimov et al. (Dimov 2022) | 2022 | – | 89.00% |
| Monsalve-Tellez et al. (Monsalve-Tellez, Torres-León, and Garcés-Gómez 2022) | 2022 | 94.29% | – |
| Fathololoumi et al. (Fathololoumi et al. 2022) | 2022 | 83.0% | – |
| Kordi et al. (Kordi and Yousefi 2022) | 2022 | 89.0% | – |
| Wang et al. (Wang et al. 2022) | 2022 | 97.72% | 80.75% |
| Li et al. (Li et al. 2022) | 2022 | 94.0% | – |
| ASEN (Proposed) | 2024 | 98.43% | 88.31 |

The higher performance metrics for ASEN can be attributed to several key contributions:

- Attention-Based Stacking Ensemble (ASEN): Unlike conventional models, ASEN integrates multiple neural networks using an adaptive attention mechanism. This enables the framework to dynamically assign importance to different classifiers, improving overall prediction accuracy and stability.
- Spectral and Spatial Feature Fusion: ASEN effectively leverages spectral fusion techniques, combining information from multiple wavelength bands and vegetation indices to capture intricate patterns in crop reflectance. This leads to improved discrimination between crop types.

- Enhanced Feature Selection: The framework prioritizes critical features using SHAP-based analysis, ensuring that only the most relevant spectral and spatial characteristics contribute to classification. This mitigates noise and redundancy, optimizing the model's learning capacity.

### 3.5. Limitations and Future Directions

Despite the strong performance of the ASEN framework, several limitations must be acknowledged to provide a more balanced perspective on its applicability and areas for improvement.

#### 3.5.1. Limitations

- The ASEN framework relies on multiple neural network models with an attention mechanism, resulting in higher computational complexity compared to conventional classifiers. While the model demonstrates strong classification accuracy, its current implementation has not been optimized for real-time inference. No formal benchmarking of inference speed or deployment feasibility on low-resource or edge devices has been conducted. This limits its immediate applicability in scenarios requiring on-device processing, such as on-farm decision support or embedded systems. Future work should include profiling model inference time and exploring lightweight variants for efficient deployment in resource-constrained environments.
- Dataset Specificity and Generalizability: The dataset used in this study was specifically curated for this research, limiting direct comparisons with other studies and reducing generalizability. The model's performance may not directly transfer to different geographic regions, soil conditions, or crop types, necessitating additional validation with more diverse datasets.
- Challenges in Distinguishing Spectrally Similar Crops: The classification accuracy for crops with similar spectral characteristics remains a challenge. Overlapping spectral signatures can lead to misclassification, particularly in monoculture farming regions. Addressing this issue may require additional feature engineering, integration of multispectral fusion techniques, or incorporating complementary data sources.
- Temporal Constraints and Seasonal Variability: This study relies on data collected exclusively during January and February 2023, which presents a limited temporal window that may not capture the full range of seasonal variations in crop development. As a result, the model's ability to generalize across different growth stages and planting cycles may be restricted.
- Spectral Similarity and Crop Misclassification: The high spectral similarity between certain crops—such as maize and wheat or potato when compared with mustard—poses a challenge for accurate classification, as reflected in the observed misclassification errors.
- Environmental and External Influences: Factors such as weather variations, soil conditions, irrigation practices, and disease outbreaks were not explicitly modeled. These external influences can significantly impact crop health and spectral responses, leading to variations in classification accuracy.

#### 3.5.2. Future Research Directions

To address these limitations and further enhance the ASEN framework, several key research directions are recommended:

- Enhancing Computational Efficiency: Future work should explore model compression techniques, such as pruning or quantization, to reduce computational overhead. The use of lightweight neural architecture and edge AI implementations could facilitate real-time or near-real-time deployment in practical agricultural settings.
- Expanding Dataset Diversity: Incorporating datasets from diverse geographic locations, soil conditions, and cropping systems will improve the robustness and generalizability of the ASEN framework. Including multi-seasonal and multi-regional data can help validate the model's effectiveness across varying agricultural landscapes.
- Integrating Multispectral and SAR Data: To improve the classification of spectrally similar crops, future studies should incorporate Synthetic Aperture Radar (SAR) data, LiDAR, or hyperspectral imaging. These additional data sources can provide complementary information beyond optical spectral reflectance, improving discrimination between crop types.
- Temporal and Multi-Season Analysis: Extending the framework to leverage multi-temporal satellite imagery can enable tracking of crop growth cycles and seasonal variations. This approach can improve predictive accuracy and enable dynamic monitoring throughout the growing season.
- Incorporating Environmental Factors: Future iterations of ASEN should integrate meteorological data, soil properties, and hydrological variables to account for external influences on crop classification. Machine learning models that fuse remote sensing and environmental datasets could enhance predictive capabilities.
- Temporal Coverage and Seasonal Variability: This study relies on data collected exclusively during January and February 2023, which presents a limited temporal window that may not capture the full range of seasonal variations in crop development. As a result, the model's ability to generalize across different growth stages and planting cycles may be restricted. Seasonal dynamics such as changes in phenology, weather conditions, and cultivation practices could significantly affect model accuracy if not represented in the training data.
- Spectral Similarity and Crop Misclassification: To address this limitation, future research could incorporate multi-temporal imagery capturing distinct phenological stages, which may enhance spectral separability. Additionally, integrating ancillary data such as soil type, crop calendars, or thermal and SAR (Synthetic Aperture Radar) imagery can provide complementary features that help disambiguate spectrally similar classes.
- Temporal Coverage and Seasonal Variability: Seasonal dynamics such as changes in phenology, weather conditions, and cultivation practices could significantly affect model accuracy if not represented in the training data. Future studies should aim to incorporate multi-temporal datasets spanning various seasons to enhance robustness, or at minimum, acknowledge this constraint when interpreting model performance.
- Developing Real-Time Monitoring Systems: Advancements in cloud-based processing and edge computing should be explored to enable real-time crop monitoring. Deploying ASEN on cloud platforms with automated data pipelines could facilitate continuous analysis for precision agriculture applications.

- Creating User-Friendly Applications: To bridge the gap between research and practical implementation, developing web-based or mobile applications with intuitive user interfaces can enable farmers and agronomists to leverage ASEN for decision-making in crop management.

## 4. Conclusion

The ASEN framework proposed in this study represents a significant advancement in crop type classification using satellite remote sensing data. By integrating multiple neural networks with an attention mechanism, ASEN effectively enhances predictive performance, achieving an accuracy of 98.43%, an F1-score of 89.02%, and an AUC of 93.42%. The attention mechanism enables the model to focus on the most relevant features, addressing challenges in distinguishing spectrally similar crops. Extensive experimental evaluation confirms ASEN's superiority over traditional classifiers such as Support Vector Machines and Logistic Regression, while comparisons with state-of-the-art models highlight its robustness and adaptability. Despite its strengths, limitations such as dataset specificity, computational complexity, and challenges in generalizing across diverse regions and seasonal variations must be addressed in future research. Enhancing dataset diversity, incorporating additional data sources like SAR imagery, and improving computational efficiency will further strengthen ASEN's scalability and real-time applicability. The ASEN establishes a new benchmark for machine learning-driven crop classification models, reinforcing the role of remote sensing technologies in precision agriculture and sustainable farming. Its high accuracy and potential for real-time monitoring make it a valuable tool for data-driven decision-making, contributing to agricultural intelligence, resource optimization and global food security efforts.

**Funding**

The study has not obtained any specific funding.

**Conflict of interest**

The authors state that there is no conflict of interest.

**Ethics approval**

Not required.

**Data availability**

The dataset is under a funding agreement and will be publicly released after the project concludes.

**Code availability**

https://github.com/nisarahmedrana/crop_classification_ASEN/edit/main/README.md

## Authors' contributions

Zeeshan Ramzan led data gathering, pre-processing, and manuscript structuring. Nisar Ahmed developed the ASEN model, implemented experiments, and optimized feature selection. Qurat-ul-Ain contributed to feature selection, experimental validation, and manuscript coherence. Shahzad Asif and Muhammad Shahbaz supervised the study, refining methodology, experimental design and reviewed the manuscript. Rabin Chakrabortty optimized spectral fusion, data analysis, and literature review. Ahmed F. Elaksher ensured statistical rigor, result interpretation, and practical implications. All authors reviewed and approved the final manuscript.

## A. Spectral Bands

Table 6 Description of Spectral bands of Landsat 7 & 8

| Bands | Wavelength (µm) | Resolution (m/pixel) | Description |
|---|---|---|---|
| Band 1 | 0.43-0.45 | 30 | Coastal aerosol band is sensitive to atmospheric conditions and vegetation properties and adept at discerning fine-scale variations in vegetation health and stress levels. |
| Band 2 | 0.45-0.51 | 30 | The blue band is crucial for early vegetation growth monitoring, especially for assessing young plants like seedlings. It can also detect specific plant diseases showing a bluish tint |
| Band 3 | 0.53-0.59 | 30 | The green band is vital for evaluating vegetation health and vigor, particularly during growth phases. It helps distinguish between healthy and stressed vegetation, enabling early stress detection. |
| Band 4 | 0.64-0.67 | 30 | The red band is vital for estimating chlorophyll content in crops. It provides insights into overall plant health and is crucial for vegetation indices like NDVI. |
| Band 5 | 0.85-0.88 | 30 | Near Infrared is highly reflective in healthy vegetation and is extensively used in vegetation indices for monitoring biomass and plant health. |
| Band 6 | 1.57-1.65 | 30 | Shortwave Infrared-1 band useful for estimating soil moisture content as it can penetrate through vegetation. This information is crucial for irrigation planning and managing water resources effectively. |
| Band 7 | 2.11-2.29 | 30 | Shortwave Infrared-2 band is valuable for mineral identification in the soil. It helps in understanding soil composition and the presence of specific minerals that affect crop growth and health. |
| Band 8 | 0.50-0.68 | 15 | The panchromatic band covers a broad range of visible light wavelengths. It provides high-resolution imagery, which is useful for detailed mapping, land use planning, and precision agriculture practices. |
| Band 9 | 1.36-1.38 | 30 | The cirrus band captures specific wavelengths that are sensitive to cirrus cloud cover. It is crucial for cloud and atmospheric monitoring and helps in identifying and mitigating the impact of clouds on imagery accuracy in agriculture research. |
| Band 10 | 10.6-11.19 | 100 | Thermal Infrared-1 captures thermal infrared radiation emitted by the Earth's surface and is used for assessing crop water stress, estimating surface temperature, and optimizing irrigation management. |
| Band 11 | 11.50-12.51 | 100 | Thermal Infrared-2 extends the thermal infrared range for more detailed temperature information. It provides even more precise temperature data, aiding in temperature-based assessments of crops and soil. |

### B. Vegetation Indices Extraction

Based on the available spectral bands and the requirements of crop classification, a comprehensive set of features is calculated in the form of vegetation indices. The selection of these vegetation indices is motivated based on the availability of spectral information and the relevance of these vegetation indices with the crop cover classification. The equations used to calculate each of these vegetation indices are provided below:

- **Normalized Difference Vegetation Index (NDVI):** The NDVI requires NIR (Band 4) and Red (Band 3) bands and can be calculated using the eq. 1:

$$NDVI = \frac{Band\ 4 - Band\ 3}{Band\ 4 + Band\ 3} \tag{1}$$

- **Enhanced Vegetation Index (EVI):** The EVI requires NIR (Band 4), Red (Band 3), and Blue (Band 1) bands and can be calculated using the eq. (*2*:

$$EVI = 2.5 \times \frac{Band\ 4 - Band\ 3}{Band\ 4 + 6 \times Band\ 3 - 7.5 \times Band\ 1 + 1} \tag{2}$$

- **Soil-Adjusted Vegetation Index (SAVI):** The SAVI requires NIR (Band 4) and Red (Band 3) bands and can be calculated using the eq. 3:

$$\mathrm{SAVI} = \left(\frac{\mathrm{Band\ 4} - \mathrm{Band\ 3}}{\mathrm{Band\ 4} + \mathrm{Band\ 3} + 0.5}\right) \times 1.5 \tag{3}$$

- **Green Normalized Difference Vegetation Index (GNDVI):** The GNDVI requires NIR (Band 4) and Green (Band 2) bands and can be calculated using the eq. 4:

$$\mathrm{GNDVI} = \frac{\mathrm{Band\ 4} - \mathrm{Band\ 2}}{\mathrm{Band\ 4} + \mathrm{Band\ 2}} \qquad 4$$

- **Red Edge Normalized Difference Vegetation Index (RENDVI): The** RENDVI requires NIR (Band 5) and Red (Band 4) bands and can be calculated using the eq. 5:

$$\mathrm{RENDVI} = \frac{\mathrm{Band\ 5} - \mathrm{Band\ 4}}{\mathrm{Band\ 5} + \mathrm{Band\ 4}} \qquad 5$$

- **Modified Soil-Adjusted Vegetation Index (MSAVI):** The MSAVI requires NIR (Band 4) and Red (Band 3) bands and can be calculated using the eq. 6:

$$\text{MSAVI} = 0.5 \times \left(2 \times \text{Band } 4 + 1 - \sqrt{(2 \times \text{Band } 4 + 1)^2 - 8 \times (\text{Band } 4 - \text{Band } 3)}\right) \quad 6$$

- **Normalized Difference Water Index (NDWI):** The NDWI requires NIR (Band 5) and Red (Band 4) bands and can be calculated using the eq. 7:

$$\text{NDWI} = \frac{\text{Band } 2 - \text{Band } 4}{\text{Band } 2 + \text{Band } 4} \quad 7$$

- **Normalized Difference Red Edge (NDRE):** The NDRE requires NIR (Band 5) and Red (Band 4) bands and can be calculated using the eq. 8:

$$\text{NDRE} = \frac{\text{Band } 5 - \text{Band } 4}{\text{Band } 5 + \text{Band } 4} \quad (8)$$

- **Simple Ratio (SR):** The SR requires NIR (Band 4) and Red (Band 3) bands and can be calculated using the eq. 9:

$$\text{SR} = \frac{\text{Band } 4}{\text{Band } 3} \quad (9)$$

- **Photochemical Reflectance Index (PRI):** The PRI requires Red (Band 3) and NIR (Band 4) bands and can be calculated using the eq. 10:

$$\text{PRI} = \frac{\text{Band } 3 - \text{Band } 4}{\text{Band } 3 + \text{Band } 4} \quad (10)$$

## C. Evaluation Metrics

### a) Classification Accuracy

Classification accuracy is fundamental in gauging the overall correctness of the model. It is computed as the ratio of correct predictions to the total number of predictions and is expressed as a percentage (Eq. 12).

$$\text{Accuracy} = \frac{\text{Number of Correct Predictions}}{\text{Total Number of Predictions}} \times 100 \quad (11)$$

### b) Precision

Precision measures the accuracy of positive predictions made by the model. It is calculated as the ratio of true positive predictions to the sum of true positive and false positive predictions (Eq. 13).

$$\text{Precision} = \frac{\text{True Positives}}{\text{True Positives} + \text{False Positives}} \quad (12)$$

### c) Recall (Sensitivity)

Recall, also known as sensitivity or true positive rate, assesses the model's ability to capture all positive instances. It is calculated as the ratio of true positive predictions to the sum of true positives and false negatives (Eq. 14).

$$\text{Recall} = \frac{\text{True Positives}}{\text{True Positives} + \text{False Negatives}} \quad (13)$$

#### d) F1-Score

F1-Score is the harmonic mean of precision and recall, providing a balanced measure that considers both false positives and false negatives (Eq. 15).

$$\text{F1-Score} = \frac{2\times\text{Precision}\times\text{Recall}}{\text{Precision}+\text{Recall}} \tag{14}$$

#### e) Area Under the Curve (AUC)

The AUC offer valuable insights into the model's performance across various classification thresholds. The ROC curve (Receiver Operating Characteristic curve) depicts the trade-off between sensitivity and specificity, while the AUC quantifies the discrimination capability, with higher values indicating better performance. In evaluating the crop cover identification model, a comprehensive strategy incorporating metrics like F1-score, accuracy, precision, recall, and AUC ensures a nuanced understanding and informs the model's readiness for practical deployment.

### Dataset Description

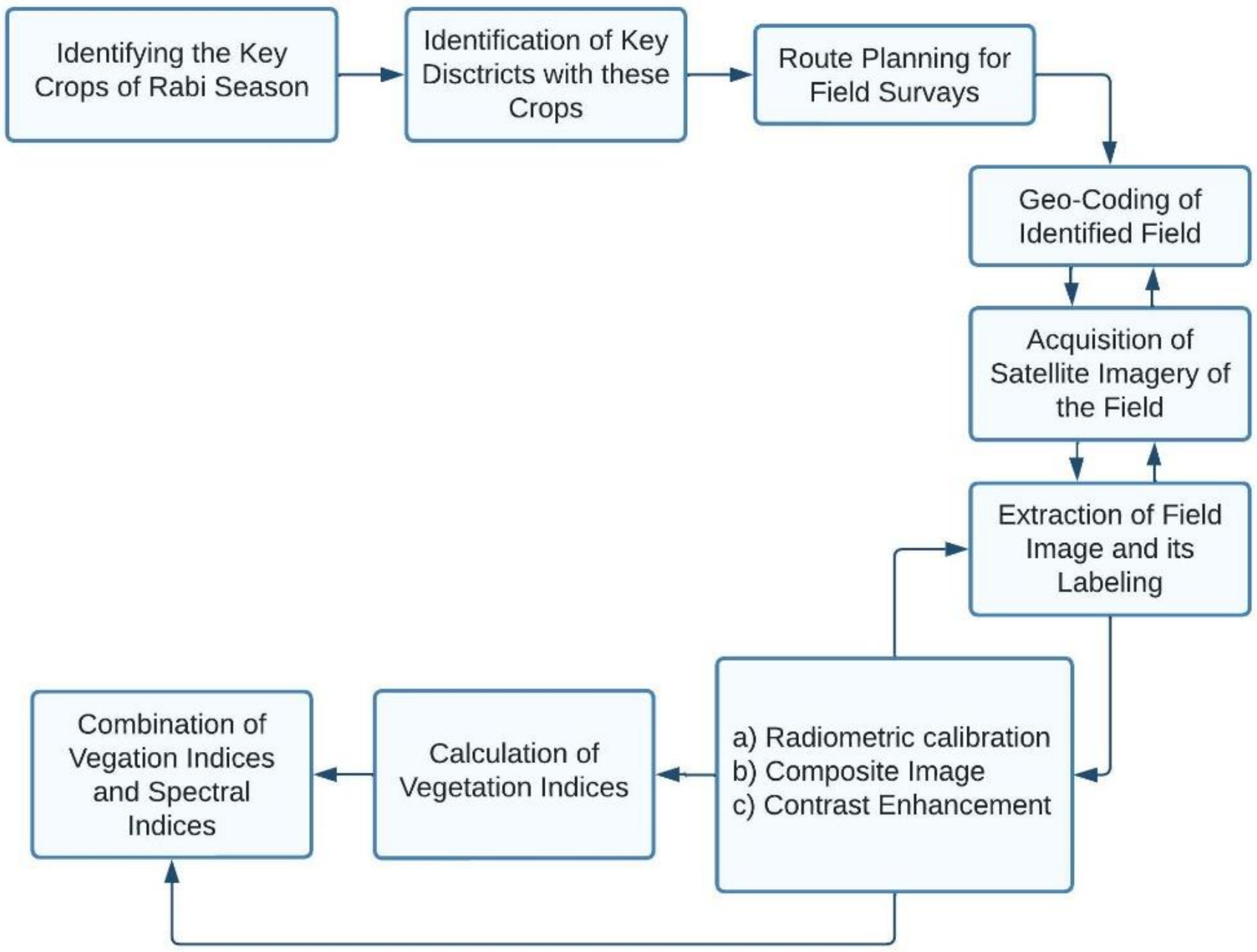

**Figure 10** Flowchart illustrating the step-by-step stages involved in preparing raw data for analysis.

## Attention-based Stacking Ensemble Network (ASEN)

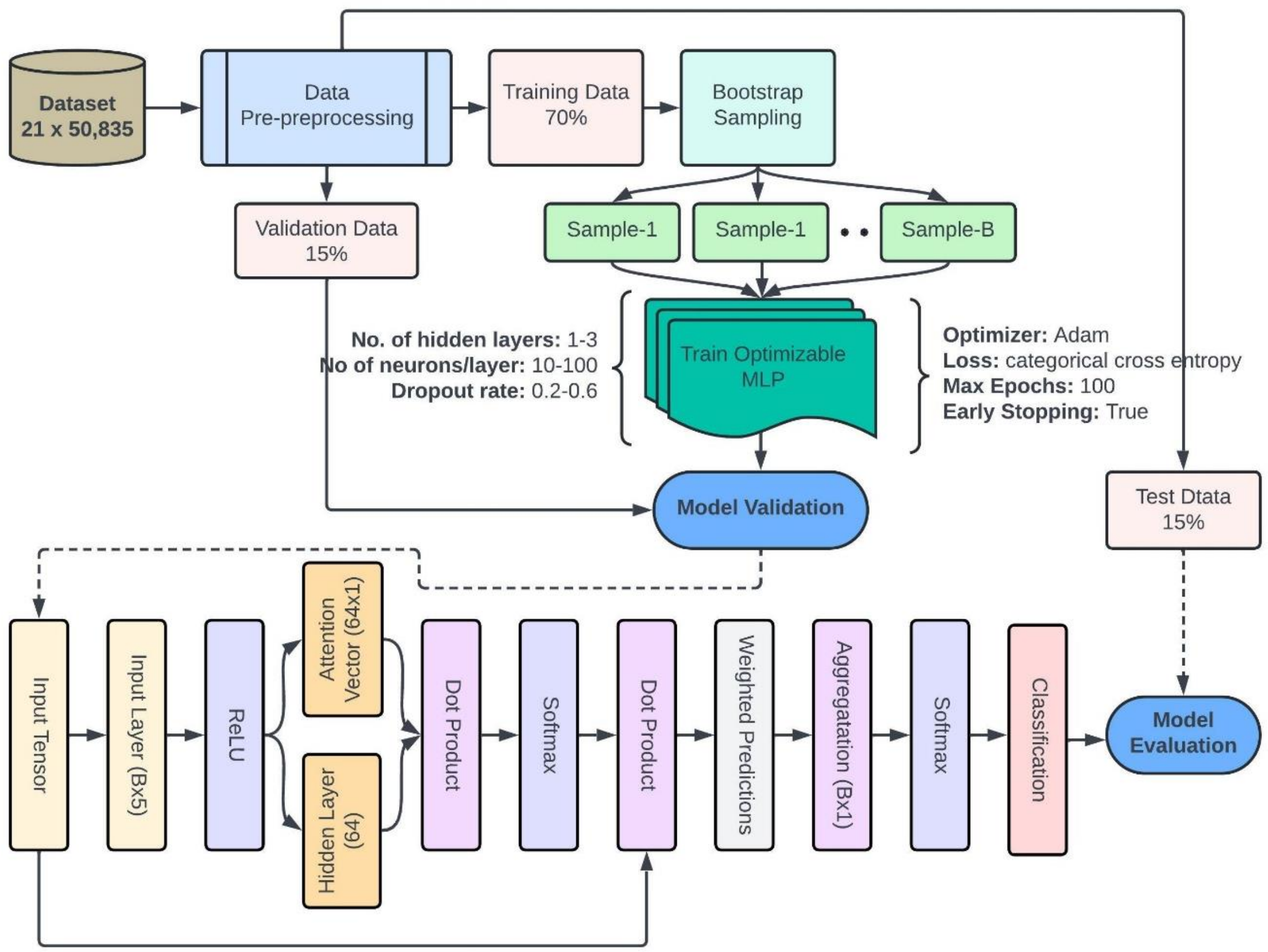

Figure 11 Schematic diagram illustrating the architecture and workflow of the ASEN model.

**Classification Map of the selected crops**

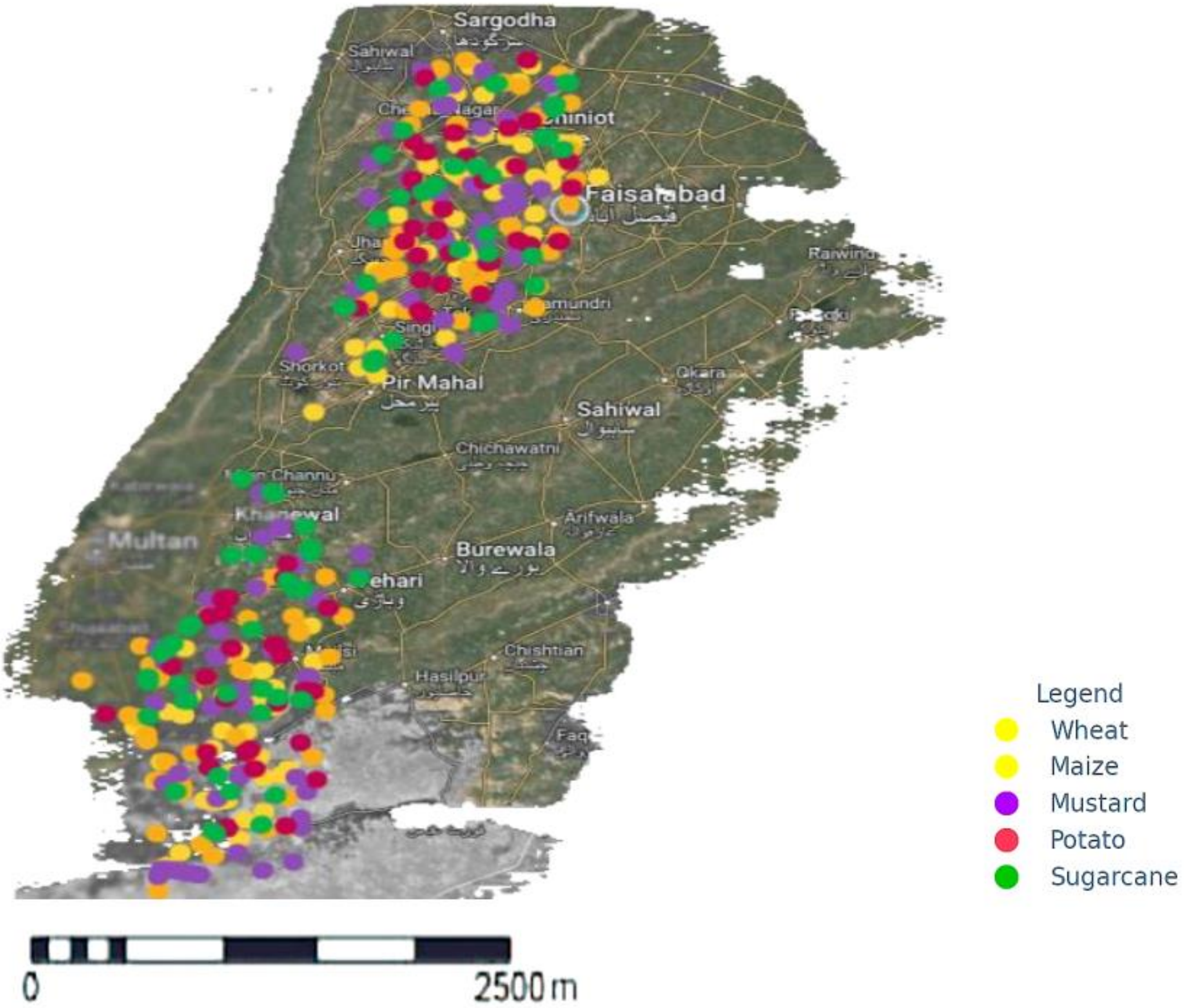

Figure 12 Visual representation of model-predicted classes for selected crops across the study area.